\documentclass[10pt,twocolumn,letterpaper]{article}
\usepackage[accsupp]{axessibility}
\usepackage{cvpr}              

\usepackage{graphicx}

\usepackage{comment}
\usepackage{amsmath,amssymb} 
\usepackage{times}
\usepackage{epsfig}
\usepackage{amsmath}
\usepackage{amssymb}
\usepackage{mathrsfs}
\usepackage{makecell}
\usepackage{booktabs}
\usepackage{multirow}
\usepackage{algorithmicx}  
\usepackage{algorithm}
\usepackage{algpseudocode}  
\usepackage{enumitem}
\usepackage{bm}
\usepackage{arydshln}
\usepackage{wrapfig}
\usepackage[subfigure]{tocloft}

\usepackage{xcolor}
\definecolor{citecolor}{HTML}{0071BC}
\definecolor{linkcolor}{HTML}{ED1C24}
\usepackage[pagebackref=false, breaklinks=true, colorlinks, citecolor=citecolor, linkcolor=linkcolor, bookmarks=false]{hyperref}

\renewcommand{\paragraph}[1]{\vspace{1.25mm}\noindent\textbf{#1}}

\usepackage[capitalize]{cleveref}
\crefname{section}{Sec.}{Secs.}
\Crefname{section}{Section}{Sections}
\Crefname{table}{Table}{Tables}
\crefname{table}{Tab.}{Tabs.}

\def\cvprPaperID{5082} 
\def\confName{CVPR}
\def\confYear{2023}

\begin{document}

\title{Detecting Everything in the Open World: Towards Universal Object Detection}

\author{Zhenyu Wang$^{1,2}$ ~~~~~~~ Yali Li$^{1,2 ~ *}$ ~~~~~~~  Xi Chen$^3$  ~~~~~~~  Ser-Nam Lim$^4$ \\ Antonio Torralba$^{5}$ ~~~~~~~ Hengshuang Zhao$^{3}$ \thanks{corresponding author} ~~~~~~~Shengjin Wang$^{1,2}$  \\ \\
$^1$ Department of Electronic Engineering, Tsinghua University\\
$^2$ Beijing National Research Center for Information Science and Technology (BNRist)\\
$^3$ The University of Hong Kong ~~~~~~ $^4$ Meta AI ~~~~~~ $^5$ Massachusetts Institute of Technology\\
{\tt\small wangzy20@mails.tsinghua.edu.cn,  \{liyali13, wgsgj\}@tsinghua.edu.cn,} \\ {\tt\small chauncey0620@gmail.com, sernamlim@fb.com, torralba@csail.mit.edu, hszhao@cs.hku.hk}
}

\maketitle

\begin{abstract}
In this paper, we formally address universal object detection, which aims to detect every scene and predict every category. The dependence on human annotations, the limited visual information, and the novel categories in the open world severely restrict the universality of traditional detectors. We propose \textbf{UniDetector}, a universal object detector that has the ability to recognize enormous categories in the open world. The critical points for the universality of UniDetector are: 1) it leverages images of multiple sources and heterogeneous label spaces for training through the alignment of image and text spaces, which guarantees sufficient information for universal representations. 2) it generalizes to the open world easily while keeping the balance between seen and unseen classes, thanks to abundant information from both vision and language modalities. 3) it further promotes the generalization ability to novel categories through our proposed decoupling training manner and probability calibration. These contributions allow UniDetector to detect over 7k categories, the largest measurable category size so far, with only about 500 classes participating in training. Our UniDetector behaves the strong zero-shot generalization ability on large-vocabulary datasets  - it surpasses the traditional supervised baselines by more than 4\% on average without seeing any corresponding images. On 13 public detection datasets with various scenes, UniDetector also achieves state-of-the-art performance with only a 3\% amount of training data.  \footnote{Codes are available at https://github.com/zhenyuw16/UniDetector.}
\end{abstract}





\vspace{-0.5cm}

\section{Introduction}


Universal object detection aims to \emph{detect everything} in every scene. Although existing object detectors \cite{ren2015faster, he2017mask, redmon2016you, lin2017focal} have made remarkable progress, they heavily rely on large-scale benchmark datasets \cite{everingham2010pascal, lin2014microsoft}. However, object detection varies in categories and scenes (\emph{i.e.}, domains). In the open world, where significant difference exists compared to existing images and unseen classes appear, one has to reconstruct the dataset again to guarantee the success of object detectors, which severely restricts their open-world generalization ability. In comparison, even a child can generalize well rapidly in new environments. As a result, universality becomes the main gap between AI and humans. Once trained, a universal object detector can directly work in unknown situations without any further re-training, thus significantly approaching the goal of making object detection systems as intelligent as humans.

\begin{figure}[t]
\centering
\setlength{\abovecaptionskip}{0pt}
\setlength{\belowcaptionskip}{0pt}
\includegraphics[width=\columnwidth]{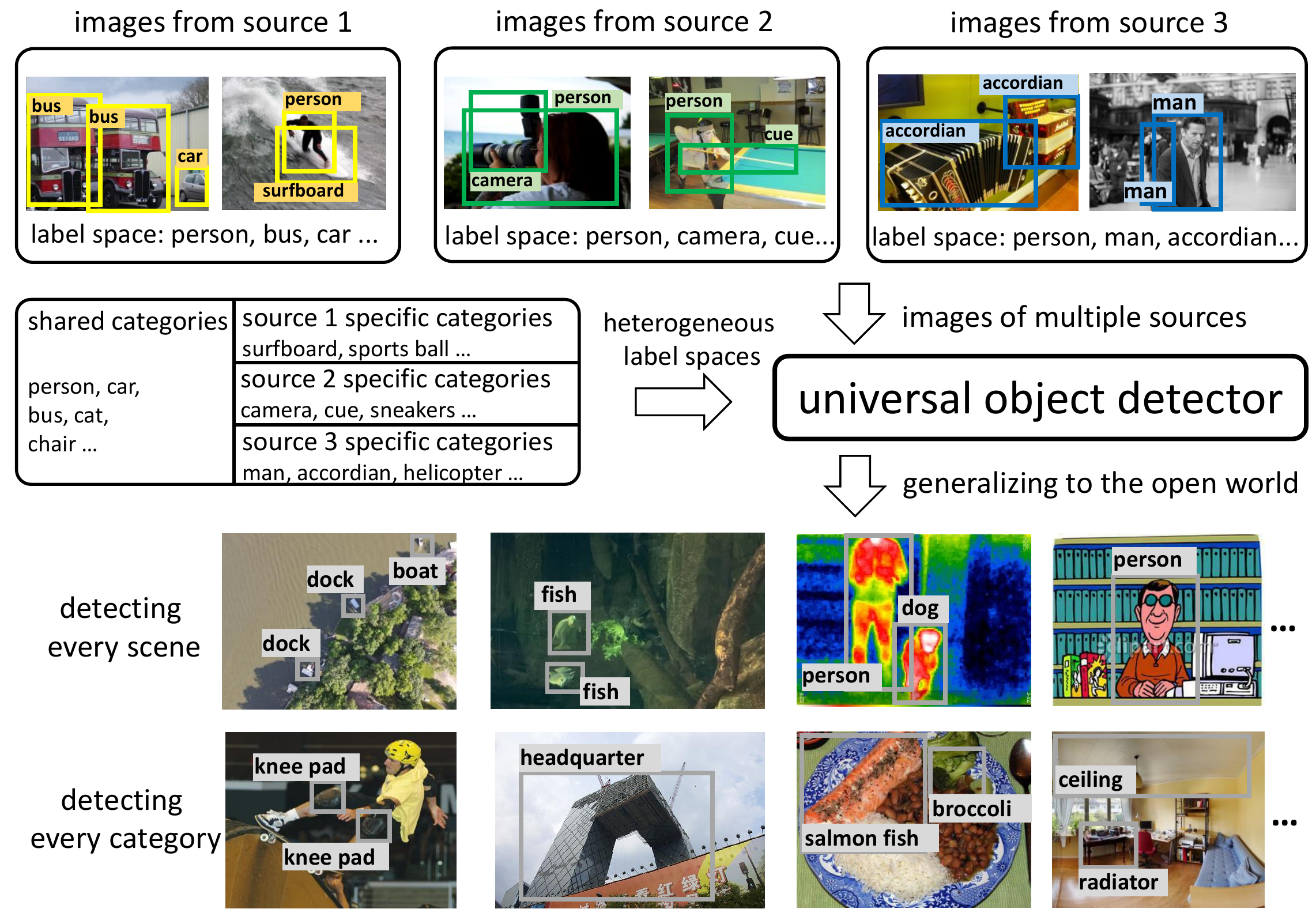}
\caption{\textbf{Illustration for the universal object detector.} It aims to detect every category in every scene and should have the ability to utilize images of multiple sources with heterogeneous label spaces for training and generalize to the open world for inference.}
\label{fig:universal}
\end{figure}

A universal object detector should have the following two abilities. First, \emph{it should utilize images of multiple sources and heterogeneous label spaces for training}. Large-scale collaborative training in classification and localization is required to guarantee that the detector can gain sufficient information for generalization. Ideal large-scale learning needs to contain diversified types of images as many as possible with high-quality bounding box annotations and large category vocabularies. However, restricted by human annotators, this cannot be achieved. In practice, unlike small vocabulary datasets \cite{everingham2010pascal, lin2014microsoft}, large vocabulary datasets  \cite{gupta2019lvis, krishna2017visual} tend to be noisily annotated, sometimes even with the inconsistency problem. In contrast, specialized datasets \cite{cordts2016cityscapes, yu2020bdd100k, zhu2018visdrone} only focus on some particular categories. To cover adequate categories and scenes, the detector needs to learn from all the above images, from multiple sources with heterogeneous label spaces, so that it can learn comprehensive and complete knowledge for universality. Second, \emph{it should generalize to the open world well}. Especially for novel classes that are not annotated during training, the detector can still predict the category tags without performance degradation. However, pure visual information cannot achieve the purpose since complete visual learning demands human annotations for fully-supervised learning.

In this paper, we formally address the task of universal object detection. To realize the above two abilities of the universal object detector, two corresponding challenges should be solved. The first one is about training with multi-source images. Images collected from different sources are associated with heterogeneous label spaces. Existing detectors are only able to predict classes from one label space, and the  dataset-specific taxonomy and annotation inconsistency among datasets make it hard to unify multiple heterogeneous label spaces. The second one is about novel category discrimination. Motivated by the recent success of image-text pre-training \cite{radford2021learning, jia2021scaling, zhai2022lit}, we leverage their pre-trained models with language embeddings for recognizing unseen categories. However, fully-supervised training makes the detector focus on categories that appear during training. At the inference time, the model will be biased towards base classes and produce under-confident predictions for novel classes. Although language embeddings make it possible to predict novel classes, the performance of them is still far less than that of base categories.

We propose UniDetector, a universal object detection framework, to address the above two problems. With the help of the language space, we first investigate possible structures to train the detector with heterogeneous label spaces and discover that the partitioned structure promotes feature sharing and avoids label conflict simultaneously. Next, to exploit the generalization ability to novel classes of the region proposal stage, we decouple the proposal generation stage and RoI classification stage instead of training them jointly. Such a training paradigm well leverages their characteristics and thus benefits the universality of the detector. Under the decoupling manner, we further present a class-agnostic localization network (CLN) for producing generalized region proposals. Finally, we propose probability calibration to de-bias the predictions. We estimate the prior probability of all categories, then adjust the predicted category distribution according to the prior probability. The calibration well improves the performance of novel classes. 

Our main contributions can be summarized as follows:

\begin{itemize}[topsep=0pt, parsep=0pt, itemsep=0pt, partopsep=0pt]
    \item We propose UniDetector, a universal detection framework that empowers us to utilize images of heterogeneous label spaces and generalize to the open world. To the best of our knowledge, this is the first work to formally address universal object detection.
    \item Considering the difference of generalization ability in recognizing novel classes, we propose to decouple the training of proposal generation and RoI classification to fully explore the category-sensitive characteristics.
    \item We propose to calibrate the produced probability, which balances the predicted category distribution and raises the self-confidence of novel categories.
\end{itemize}

Extensive experiments demonstrate the strong universality of UniDetector. It recognizes the most measurable categories. Without seeing any image from the training set, our UniDetector achieves a 4\% higher AP on existing large-vocabulary datasets than fully-supervised methods. Besides the open-world task, our UniDetector achieves state-of-the-art results in the closed world - 49.3\% AP on COCO with a pure CNN model, ResNet50, and the 1$\times$ schedule.

\section{Related Work}

\paragraph{Object detection} aims to predict category tags and bounding box coordinates of each object within an image. Existing methods can be generally divided into two-stage and one-stage methods. Two-stage detectors mainly include RCNN \cite{girshick2014rich} and its variants \cite{girshick2015fast, ren2015faster, he2017mask, Cai_2018_CVPR}. They usually extract a series of region proposals first, then perform classification and regression. In comparison, one-stage detectors \cite{redmon2016you, liu2016ssd, lin2017focal} directly generate classification results for the anchors. Different from these methods, models such as \cite{law2018cornernet, zhou2019objects, tian2019fcos, zhang2020bridging} are anchor-free for object detection. Recently, transformer-based methods \cite{carion2020end, zhu2020deformable, dai2021up, li2022dn, zhang2022dino} also develop rapidly. However, most of these methods can only work in the closed world. 

\begin{figure*}[t]
\centering
\setlength{\abovecaptionskip}{0pt}
\setlength{\belowcaptionskip}{0pt}
\includegraphics[width=\textwidth]{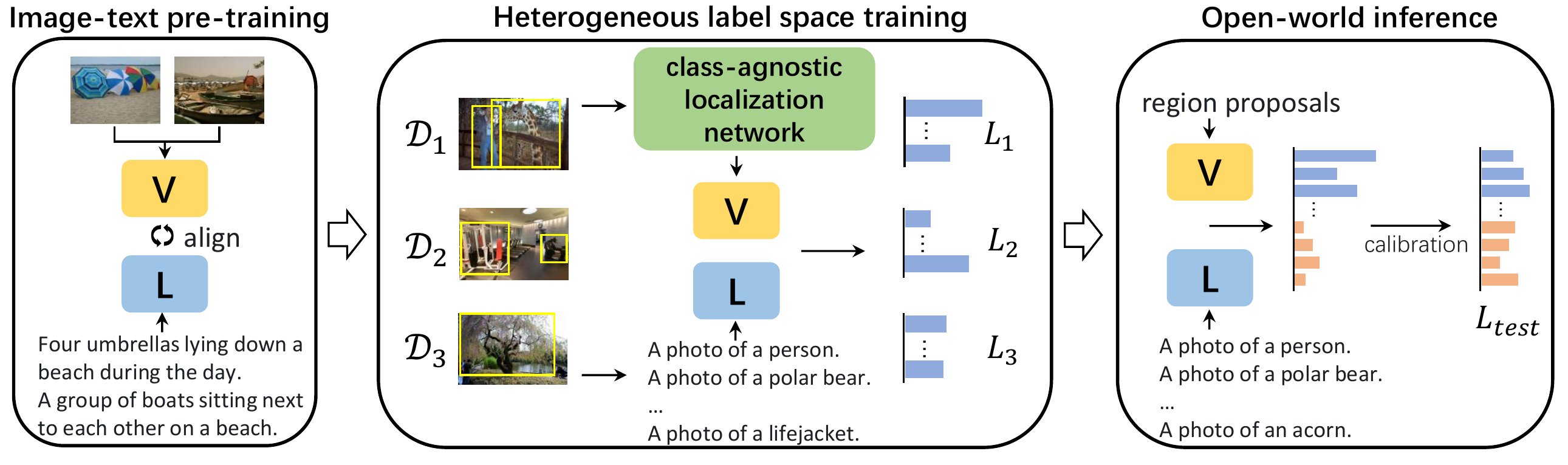}
\caption{\textbf{Overview of UniDetector.} It consists of three steps. With the image-text pre-training parameters, UniDetector is trained with images of different sources with multiple label spaces. In this way, it can directly detect in the open world for inference. 'V' denotes the module to process visual information, and 'L' denotes the language encoder. The first stage conducts image-text pre-training to align the two spaces, the second stage trains with images of heterogeneous label spaces in the decoupling manner, and the third stage applies probability calibration to maintain the
balance.}
\label{fig:frame}
\end{figure*}

\paragraph{Open-vocabulary object detection.} Traditional object detection can only detect categories that appear at the training time. In universal object detection, categories that need to be detected cannot be known in advance. For this purpose, zero-shot object detection \cite{bansal2018zero, zhu2019zero, zhu2020don, rahman2020improved} is proposed and aims to generalize from seen to unseen categories. However, their performance is still far behind fully-supervised methods. Based on these researches, open-vocabulary object detection \cite{zareian2021open} forwards the task. By involving image-text aligned training, the unbounded vocabularies from the texts benefit the generalization ability of the model for detecting novel categories. With the development of large-scale image-text pre-training works \cite{radford2021learning, jia2021scaling, zhai2022lit}, recent methods \cite{gu2021open, du2022learning, zhong2022regionclip, minderer2022simple, feng2022promptdet} have considered adopting such pre-trained parameters in open-vocabulary detection, and boosting the performance and category vocabulary to a large extent. Despite their success, existing methods still target at transferring within a single dataset. Besides, their seen categories are usually more than unseen categories. Their generalization ability is thus restricted.

\paragraph{Multi-dataset object detection training.} Previous object detection methods focus on only one single dataset. Since only one dataset is involved in training, both the dataset scale and vocabulary are limited. Recently, training on multiple datasets \cite{cai2022bigdetection, shi2021multi} has been used to boost the model's robustness and expand the detector's vocabulary size. The difficulty of multi-dataset training for object detection is to utilize multiple heterogeneous label spaces. For this purpose, \cite{zhao2020object} leverages pseudo labels to unify different label spaces, \cite{wang2019towards, zhou2022simple} adopt a partitioned structure, and \cite{meng2022detection} utilizes language embeddings. However, these methods still focus on detecting in the closed world. Different from them, we target generalizing in the open world.

\section{Preliminary}

Given an image $I$, object detection aims to predict its label $y=\{(b_i, c_i)\}_{i=1}^m$, which consists of bounding box coordinates $b_i$ and category label $c_i$. We are usually given a single dataset $\mathcal{D}_{train}=\{(I_1, y_1), ..., (I_n, y_n)\}$ and the goal is to inference on the test dataset $\mathcal{D}_{test}$. 

Traditional object detection can only work in the closed world, where images are restricted to a single dataset. The dataset has its own label space $L$. Each object category $c_i$ from either $\mathcal{D}_{train}$ or $\mathcal{D}_{test}$ belongs to the same predefined label space (\emph{i.e.} class vocabulary) $L$.

In this work, we propose a brand new object detection task, which focuses on the universality of detectors. At the training time, we utilize images from multiple sources. That is, images of heterogeneous label spaces $L_1, L_2, ... L_n$. At the inference time, the detector predicts class labels from a new label space $L_{test}$, which is provided by the users. 


The advances in traditional object detection cannot be trivially adapted to our universal detection task. The main reason is that there exist novel categories at the inference time: $L_{novel} = L_{test} \backslash \bigcup_{i=1}^n L_i $. Techniques in traditional object detection benefit base categories $L_{base} = \bigcup_{i=1}^n L_i$ but may hurt novel categories. The core issue of our work is therefore how to utilize images of heterogeneous label spaces, and how to generalize to novel categories.


\begin{figure*}[t]
\centering
\setlength{\abovecaptionskip}{0pt}
\setlength{\belowcaptionskip}{0pt}
\begin{subfigure}{0.36\textwidth}
\centering
\includegraphics[height=3.3cm]{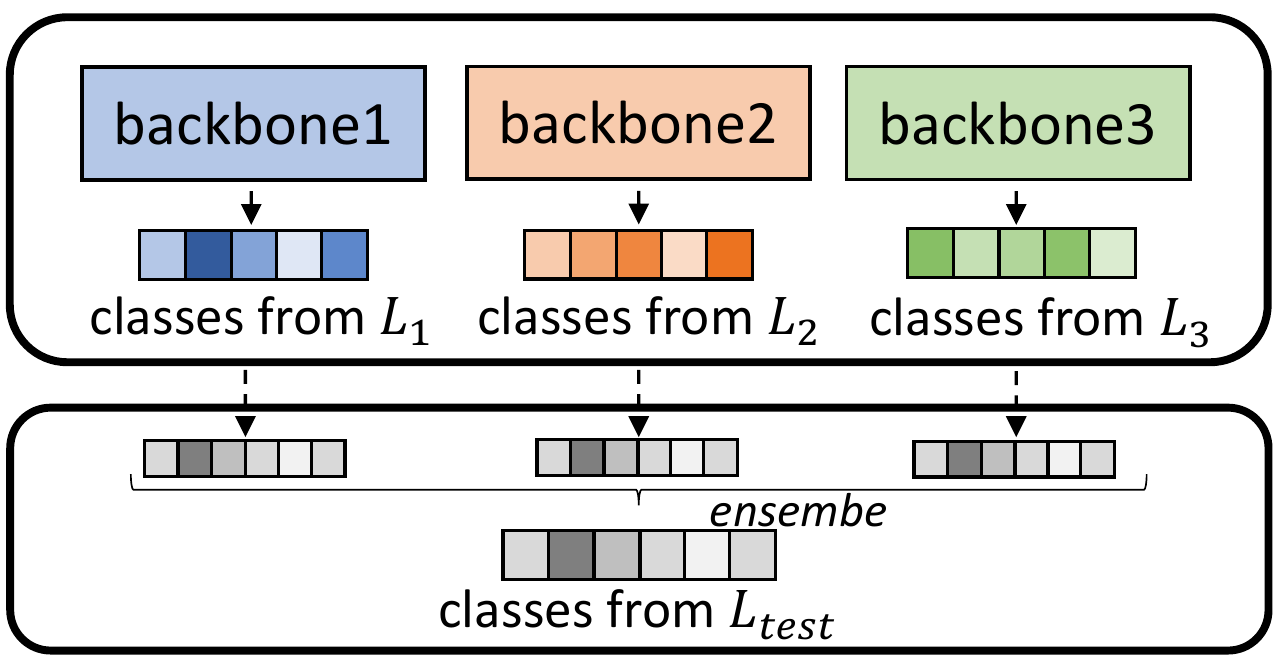}
\caption{seperate label spaces}
\label{fig:m1}
\end{subfigure}
\begin{subfigure}{0.3\textwidth}
\centering
\includegraphics[height=3.3cm]{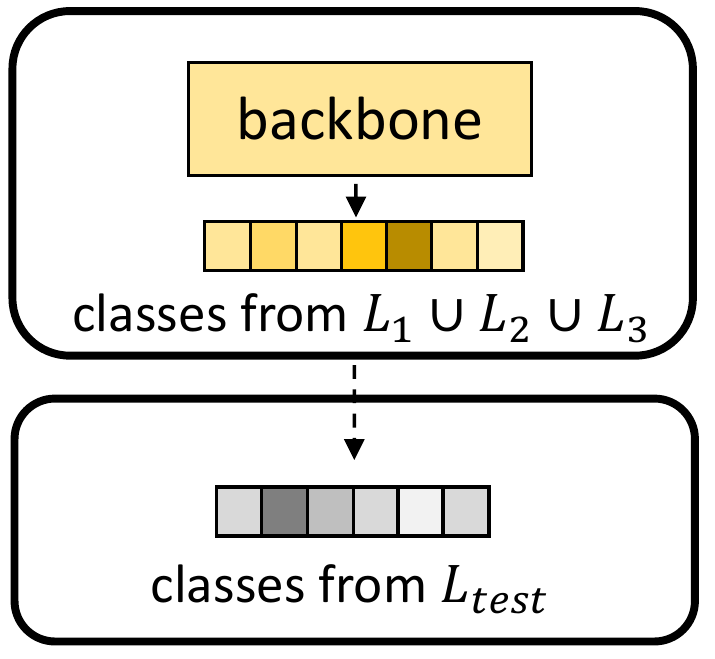} 
\caption{unified label space}
\label{fig:m2}
\end{subfigure}
\begin{subfigure}{0.33\textwidth}
\centering
\includegraphics[height=3.3cm]{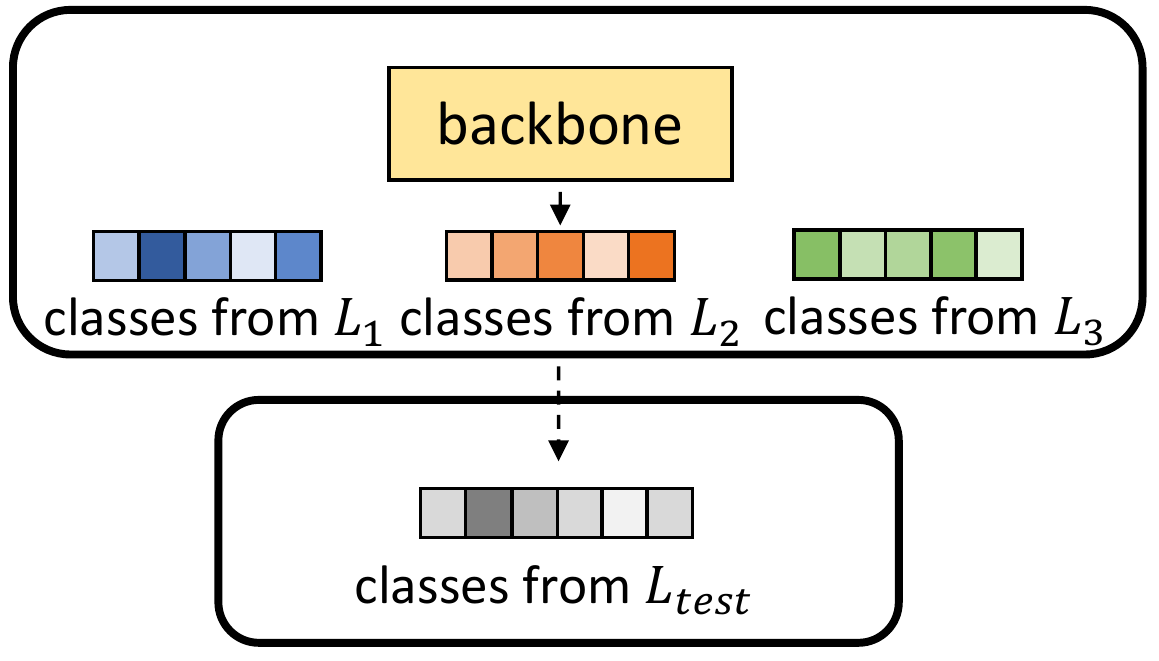} 
\caption{partitioned label space}
\label{fig:m3}
\end{subfigure}
\caption{\textbf{Possible structures to utilize images from heterogeneous  label spaces for training.} The above boxes denote the structure for training, and the below boxes denote the inference process. All the classification head here adopts the similarity between the region features and language embeddings. The separate structure trains individual networks and ensembles them for inference, the unified structure unifies the multiple datasets into one dataset, and the partitioned structure shares the same backbone but different classification heads. }
\label{fig:multi}
\end{figure*}

\section{The UniDetector Framework}

We propose the UniDetector framework to address the universal object detection task, which is illustrated in Fig. \ref{fig:frame}. The basic process consists of three steps.

\paragraph{Step1: Large-scale image-text aligned pre-training.} Traditional fully-supervised learning with only visual information relies on human annotations, which restricts the universality. Considering the generalization ability of language features, we introduce language embeddings to assist detection. Inspired by the recent success of language-image pre-training, we involve the embeddings from a pre-trained image-text model \cite{radford2021learning, jia2021scaling, zhong2022regionclip, zhai2022lit}. We adopt RegionCLIP \cite{zhong2022regionclip} pre-trained parameters for our experiments.


\paragraph{Step2: Heterogeneous label space training.} Unlike traditional object detection, which concentrates on a single dataset with the same label space, we collect images from different sources with heterogeneous label spaces to train the detector. The various training images are necessary for the detector's universality. Meanwhile, we adopt the decoupling manner during the training process instead of the previous joint training. 

\paragraph{Step3: Open-world inference.} With the trained object detector and the language embeddings from the test vocabulary, we can perform detection in the open world directly for inference without any finetuning. However, since novel categories do not appear during training, the detector is easy to generate under-confident predictions. We propose probability calibration to keep the inference balanced between base categories and novel categories in this step.

\subsection{Heterogeneous Label Space Training}
Existing object detectors can only learn from images with one label space because of their single classification layer. To train with heterogeneous label spaces and obtain sufficiently diversified information for universality, we present three possible model structures, as shown in Fig. \ref{fig:multi}.


One possible structure is to train with separate label spaces. As in Fig. \ref{fig:m1}, we train multiple models on every single dataset (\emph{i.e.}, label space). With new language embeddings at the inference time, each single model can perform inference on the test data. These individual test results can be combined to obtain the ultimate detection boxes. Another structure is to unify multiple label spaces into one label space, as in Fig. \ref{fig:m2}. Then we can treat these data the same as before. Since images are treated as if they are from one single dataset, they can be processed with techniques like Mosaic \cite{bochkovskiy2020yolov4} or Mixup \cite{zhang2017mixup} to boost information integration among different label spaces. With the help of language embeddings for classification, we can also use the partitioned structure in Fig. \ref{fig:m3}, where images of multiple sources share the same feature extractor but have their own classification layer. At the inference time, we can directly use the class embedding of test labels to avoid label conflict.

We then need to consider the data sampler and the loss function. When the data turn large-scale, an unavoidable problem is their long-tailed distribution \cite{shao2019objects365, kuznetsova2020open, krishna2017visual}. Samplers like the class-aware sampler (CAS) \cite{peng2020large} and the repeat factor sampler (RFS) \cite{gupta2019lvis} are helpful strategies for multi-dataset detection in the closed world \cite{zhou2022simple}. However, the open-world performance is unaffected. The reason is that the core issue here is about novel classes. With language embeddings, the adverse effect of the long-tailed problem becomes negligible. We thus adopt the random sampler.

Likewise, loss functions like equalized loss \cite{tan2020equalization, tan2021equalization} and seesaw loss \cite{wang2021seesaw} influence universal object detection little. Instead, the sigmoid-based loss is more suitable since the classification of base and novel categories will not interfere with each other under the sigmoid function. To avoid an excessive value of sigmoid-based classification loss when the number of categories increases, we randomly sample a certain number of categories as negative ones.

\paragraph{Decoupling proposal generation and RoI classification.} A two-stage object detector consists of a visual backbone encoder, a RPN and a RoI classification module. Given an image $I$ from the dataset $D$ with the label space $L$, the network can be summarized as: $\{z_{ij}\}_{j=1}^{|L|} = \Phi_{RoI} \circ \Phi_{RPN} \circ \Phi_{backbone}$, $p_{ij} = 1/(1 + {\rm exp}(-z_{ij}^Te_j/\tau)), j \in L$,
where $p_{ij}$ is the probability of the $i$-th region for the category $j$, $\{z_{ij}\}_{j=1}^{|L|}$ denotes the logit outputs from the RoI head, and $e_j$ is the language embedding of the category $j$.

The region proposal generation stage and the RoI classification stage act differently when it comes to universal detection. The proposal generation stage maintains satisfying universality ability since its class-agnostic classification can be easily extended to novel classes. In contrast, the class-specific RoI classification stage cannot even work for novel categories. Even with language embeddings, it is still biased to base classes. The distinct properties affect their joint training since the sensitivity of the classification stage to novel classes hampers the universality ability of the proposal generation stage. Consequently, we decouple these two stages and train them separately to avoid such conflict.

Specifically, the region proposal generation stage is initialized with traditional ImageNet pre-trained parameters and trained in a class-agnostic way. After training, it produces a series of region proposals. With the generated proposals, the RoI classification stage is trained in the Fast RCNN \cite{girshick2015fast} manner. This stage is initialized with image-text pre-trained parameters for predicting unseen categories. These two kinds of pre-trained parameters also contain complementary features, which provide more comprehensive information for universal object detection.

\begin{figure}[t]
\centering
\setlength{\abovecaptionskip}{0pt}
\setlength{\belowcaptionskip}{0pt}
\includegraphics[width=\columnwidth]{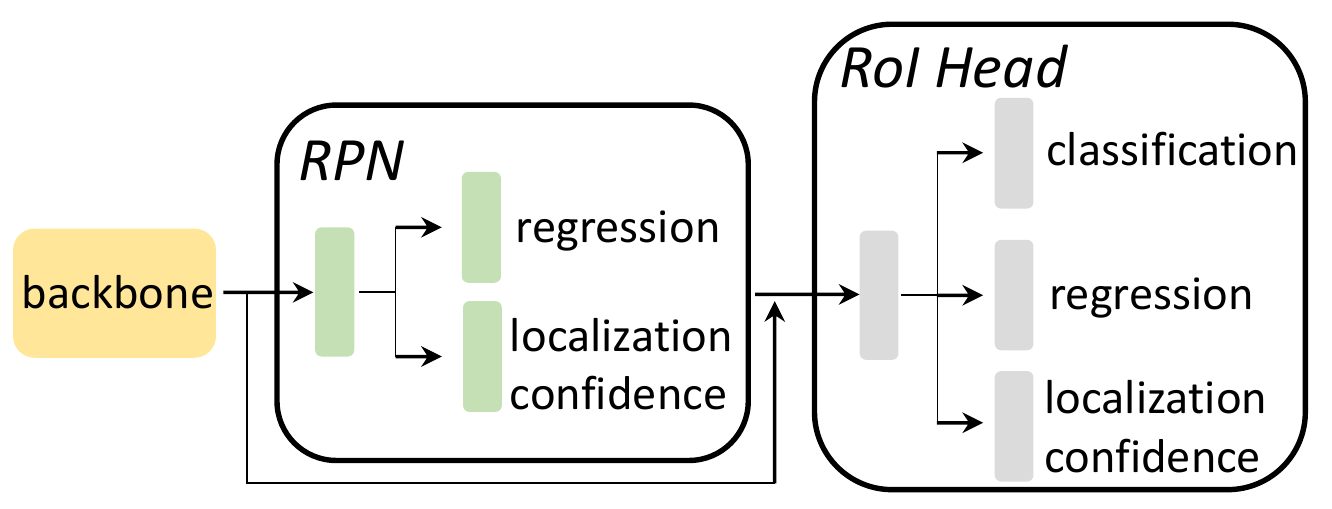}
\caption{\textbf{Illustration of class-agnostic localization network.} The localization confidence and class-agnostic classification contribute together to generating proposals for universal detection.}
\label{fig:cln}
\end{figure}

\paragraph{Class-agnostic localization network.}
To produce generalized proposals in the open world, we present the class-agnostic localization network (CLN), as illustrated in Fig. \ref{fig:cln}. Instead of a single RPN module, our CLN contains both the RPN and the RoI head to generate proposals for universal object detection. Such kind of network promotes box refinement during proposal generation. We mainly adopt localization-based objectness for object discovery since localization-related metric tends to be robust to novel objects in the open world \cite{kim2022learning}. In the RoI head, based on the localization confidence, we also keep the binary class-agnostic classification because it provides strong supervised signals to the network training. For the $i$-th proposal, denoting its localization confidence from the RPN as $s_i^{r_1}$, from the RoI head as $s_i^{r_2}$, and its classification confidence as $s_i^c$, the ultimate confidence from our CLN can be obtained through geometric weighting: $\eta_i = (s_i^c) ^\alpha \cdot (s_i^{r_1}s_i^{r_2}) ^{(1-\alpha)}$, where $\alpha$ is a pre-defined hyperparameter.

\subsection{Open-world Inference}
With the language embeddings of the test vocabulary $L_{test}$, our trained detector can directly perform inference in the open world. However, since only base categories appear during training, the trained detector will be biased toward base categories. As a result, boxes of base categories in detection results tend to have a larger confidence score than novel categories, thus predominating in the inference process. Considering the huge number of novel categories, the over-confidence of base classes will easily make the detector ignore novel category instances of a larger number and hurt the detector's performance in the open world. 

To avoid the bias issue, we propose probability calibration for post-processing the predictions. The purpose of calibration is to decrease the probability of base categories and increase that of novel categories, thus balancing the ultimate probability prediction. The probability calibration is illustrated as follows:
\begin{equation}
    p_{ij} = \frac{1}{1 + {\rm exp}(-z_{ij}^Te_j/\tau)} / \pi_j^\gamma, j \in L_{test}
    \label{equ:calibration}
\end{equation}


Our probability calibration is mainly about dividing the original probability with a prior probability $\pi_j$ of the category $j$. The prior probability $\pi_j$ records the bias of the network to category $j$. $\gamma$ is a pre-defined hyperparameter. A larger $\pi_j$ indicates that the model is more biased toward the category. After calibration, its probability turns smaller, which contributes to the probability balance. We can conduct inference on the test data first and use the number of categories within the results to obtain $\pi_j$. If the number of test images is too small to estimate the accurate prior probability, we can also use the training images to calculate $\pi_j$.

$p_{ij}$ from Eq. \ref{equ:calibration} reflects the class-specific prediction for the $i$-th region proposal. Considering the open world generalization ability of the class-agnostic task, we multiply $p_{ij}$ with its objectness score $\eta_i$ from CLN for the detection score. After further introducing a hyper-parameter $\beta$, the final detection score is $s_{ij} = p_{ij}^\beta \eta_i^{(1-\beta)}$.


\section{Experiments}

To demonstrate the universality of our UniDetector, we conduct experiments and evaluate our UniDetector in the open world, in the traditional closed world and in the wild. Its superior performance under various conditions well illustrate its universality. 


\begin{table*}[t]
\centering
\setlength{\abovecaptionskip}{0pt}
\setlength{\belowcaptionskip}{0pt}
\caption{\textbf{The performance of UniDetector in the open world.} We evaluate it on LVIS, ImageNetBoxes and VisualGenome. S, U, P denote treating heterogeneous label spaces as separate spaces, a unified one or a partitioned one. The Faster RCNN (closed world) row is from the traditional supervised Faster RCNN C4 baseline trained on the corresponding dataset with the same random data sampler. We select 35k, 60k, 78k images from COCO, Objects365 and OpenImages respectively for training.}
\resizebox{\textwidth}{!}{
\begin{tabular}{cc|cccc|cccc|ccc|ccc}
\Xhline{1.1pt} 
\multirow{2}{*}{Training data} & \multirow{2}{*}{Structure} & \multicolumn{4}{c|}{LVIS v0.5 (1,230)} & \multicolumn{4}{c|}{LVIS v1 (1,203)} & \multicolumn{3}{c|}{ImageNetBoxes (3,622)} & \multicolumn{3}{c}{VisualGenome (7,605)}\\
 &  & AP & AP$_r$ & AP$_c$ & AP$_f$ & AP & AP$_r$ & AP$_c$ & AP$_f$ & AP & AP$_{50}$ & Loc. Acc  & AR$_{1}$ & AR$_{10}$ & AR$_{100}$\\
\hline
\multicolumn{2}{c|}{Faster RCNN (closed world)}  & 17.7 & 1.9 & 16.5 & 25.4 & 16.2 & 0.9 & 13.1& 26.4& 3.9 & 6.1 & 15.3  & 3.5 & 4.3 & 4.3\\
\hline
COCO & - & 16.4 & 18.7 & 17.1 & 14.5 & 13.7 & 13.5 & 13.6 & 13.9 & 4.8 & 6.8 & 8.3 & 4.3 & 5.9 & 5.9\\
O365 & - & 20.2 & 21.3  & 20.2 & 19.8  & 16.8 & 14.7 & 16.2 & 18.3 & 3.8 & 5.5 & 8.4  & 5.4 & 7.3& 7.3\\
OImg & - & 16.8 & 21.8 & 17.6 & 13.8 & 13.9 & 14.7 & 14.2 & 13.2 & 7.9 & 10.8 & 16.0  & 5.9 & 8.1 & 8.2\\
\hline
COCO + O365 & S & 21.0 & 22.2 & 21.8 & 19.4 & 17.5 & 16.0 & 17.2 & 18.4 & 4.5 & 6.5 & 8.9 & 6.2 & 8.5 & 8.6\\
COCO + O365 & U & 20.9 & 19.6 & 21.0 & 21.3 & 17.6 & 14.6 & 17.0 & 19.6 & 3.6 & 5.1 & 8.0 & 5.3 & 7.1 & 7.2\\
COCO + O365 (+mosaic) & U  & 21.4 & 22.3 & 21.5 & 21.0 & 16.8 & 13.5 & 16.2 & 18.9 &3.6 & 5.1& 7.7 & 5.0 & 6.8 & 6.9\\
COCO + O365 (+pseudo \cite{zhao2020object}) & U & 20.8 & 22.5 & 22.7 & 19.7 & 16.6 & 13.4 & 16.1 & 18.7 & 3.6 & 5.1 & 7.6 & 5.0 & 6.6 & 6.7\\
\hline
COCO + O365 & P  & 22.2 & 23.7 & 22.5 & 21.2 & 18.2 & 15.5 & 17.6 & 20.1 & 4.7 & 6.6 & 10.1 & 5.9 & 8.0 & 8.1\\
COCO + OImg & P  & 19.9 & 22.1 & 20.7 & 17.9 & 16.8 & 16.0 & 16.8 & 17.1 & 6.9 & 9.5 & 14.7  & 5.7 & 7.7 & 7.8\\
COCO + O365 + OImg & P  & \textbf{23.5} & \textbf{23.6} & \textbf{24.3} & \textbf{22.4} & \textbf{19.8} & \textbf{18.0} & \textbf{19.2} & \textbf{21.2} & \textbf{8.2} & \textbf{11.4} & \textbf{16.9}  & \textbf{6.5} & \textbf{8.7} & \textbf{8.8}\\
\Xhline{1.1pt} 
\end{tabular}
}
\label{tab:openworld}
\end{table*}

\noindent \textbf{Datasets.} To simulate images of multiple sources and heterogeneous label spaces, we adopt three popular object detection datasets for training the detector: COCO \cite{lin2014microsoft}, Objects365 \cite{shao2019objects365}, and OpenImages \cite{kuznetsova2020open}. COCO contains dense and high-quality annotations from human labor on 80 common classes. Objects365 is larger-scale and contains 365 classes. OpenImages consists of more images and 500 categories, and many annotations are sparse and dirty. Due to the large scale of these datasets, we randomly sample 35k, 60k, and 78k images from them respectively for training. Without specification, we all use the selected subset.

We mainly perform inference on the LVIS \cite{gupta2019lvis}, ImageNetBoxes \cite{NIPS2012_c399862d}, and VisualGenome \cite{krishna2017visual} datasets to evaluate the detector's open-world performance. Considering their large category numbers, these datasets can simulate the open-world environment to some extent. LVIS v0.5 consists of 1,230 categories, and LVIS v1 contains 1,203 categories, with 5,000 images and 19,809 images for the validation set, respectively. ImageNetBoxes contains over 3,000 categories. We random sample 20,000 images from the dataset for evaluation. To compare with the supervised closed world baseline, we sample 90,000 images as the training set. The most recent version of the VisualGenome dataset contains 7,605 categories. However, since a large number of its annotations come from machines, the annotations are pretty noisy. We select 5,000 images that do not appear in the training images for inference. 

\noindent \textbf{Evaluation metrics.} We mainly adopt the standard box AP for evaluating the performance. For the LVIS dataset, we also evaluate the performance on its rare, common and frequent categories separately, denoted as AP$_r$, AP$_c$, and AP$_f$. For the ImageNetBoxes dataset, since most of the images within it are object-centric, besides the AP and AP$_{50}$ metrics, we also adopt the top-1 localization accuracy (denoted as Loc. Acc.) from the ImageNet challenge \cite{russakovsky2015imagenet} to evaluate the object-centric classification ability of the detector. For the VisualGenome dataset, considering the noise and inconsistency of its annotations, we adopt the Average Recall (AR) metric for evaluation.

\noindent \textbf{Implementation details.} We implement our method with mmdetection \cite{chen2019mmdetection}. Without otherwise specified, we choose ResNet50-C4 \cite{he2016deep} based Faster RCNN \cite{ren2015faster} as our detector, initialized with RegionCLIP \cite{zhong2022regionclip} pre-trained parameters. All the models are trained in the 1$\times$ schedule, which is 12 epochs. For hyperparameters, $\tau$ is set to 0.01, $\gamma$ is set to 0.6, and $\alpha, \beta$ are set to 0.3 both.

\subsection{Object Detection in the Open World}

We list the open-world results of UniDetector in Tab. \ref{tab:openworld}. For comparison, we conduct supervised closed-world experiments with the same Faster RCNN C4 structure and the random data sampler. On the LVIS v0.5 dataset, the traditional supervised detector obtains the 17.7\% AP. In comparison, our UniDetector with only 35k COCO images obtains the 16.4\% AP. With only 60k Objects365 images, it obtains 20.2\% AP. With significantly fewer images and annotated categories, the detection AP is even higher. The effectiveness of our UniDetector is demonstrated: it can achieve comparable or even better performance compared to the corresponding closed-world detector, while the required training budget is less. Another noticeable result is that the traditional closed-world detector suffers from the long-tailed problem - AP$_r$ is only 1.9\% compared to the 25.4\% AP$_f$. In comparison, the AP$_r$ and AP$_f$ from our detector are significantly more balanced. This illustrates that UniDetector also greatly alleviates the long-tailed effect.

We then analyze the effects of different structures on the COCO and Objects365 datasets. We use WBF \cite{solovyev2021weighted} to ensemble two detectors for the separate label spaces. Images of different sources cannot interact during training under this structure, which restricts the feature extraction ability. For the unified space, the inconsistency labels of different datasets lead to a serious missing annotation problem. Although we adopt pseudo labels according to \cite{zhao2020object} and boost the image fusion through mosaic, the open-world AP is still not improved. In contrast, with the partitioned structure, all kinds of images train the backbone together, thus promoting feature extraction. In the classification time, the partitioned label space mitigates the label conflict. Therefore, the partitioned structure performs the best among them.

With the partitioned structure, COCO and Objects365 joint training achieves the 22.2\% AP, higher than the single results of 16.4\% and 20.2\%. We also notice that OpenImages single training obtains the 16.8\% LVIS AP, only slightly higher than COCO and even lower than Objects365. Considering the more images and categories within it, the limited performance can be attributed to its noisy annotations. However, if we further add OpenImages images to COCO and Objects365, the LVIS v0.5 AP can be improved to 23.5\%. At this time, COCO and Objects365 images have high-quality annotations, while OpenImages provides more categories but noisy annotations. Images from multiple sources cooperate and bring various information, thus contributing to better open-world performance. This is the most significant superiority of training with heterogeneous label spaces to universal object detection. A similar trend of results is also observed for LVIS v1.

We further evaluate our UniDetector on ImageNetBoxes and VisualGenome datasets. These two datasets contain more categories, thus better simulating the open-world environment. Our UniDetector keeps an excellent open-world generalization ability. On the ImageNetBoxes dataset, it obtains the 8.2\% AP, surpassing the 3.9\% AP from traditional detectors with comparable training images. It is also worth mentioning that the domain gap between the ImageNetBoxes dataset and COCO-style datasets is relatively large, since ImageNetBoxes images are mainly object-centric. In this situation, our UniDetector still generalizes well, which validates the universality of our UniDetector. On the VisualGenome dataset, where category numbers are more than 7,000, our UniDetector also obtains a higher detection result compared to the traditional Faster RCNN. The most significant improvement comes from the AR$_{100}$ metric, which is more than 4\%. Through this experiment, the category recognition ability of our UniDetector is revealed.

\subsection{Object Detection in the Closed World}

\begin{table}[t]
\centering
\setlength{\abovecaptionskip}{0pt}
\setlength{\belowcaptionskip}{0pt}
\caption{\textbf{The performance of UniDetector in the closed world.} The models are trained on COCO train2017 set with the 1$\times$ schedule (12 epochs), ResNet50 backbone and evaluated on COCO val2017. $^\S$: the method uses extra images and more epochs.}
\resizebox{\columnwidth}{!}{
\begin{tabular}{c|ccccc}
\Xhline{1.1pt} 
 Model  & AP & AP$_{50}$  & AP$_S$ & AP$_M$ & AP$_L$\\
\hline
\multicolumn{6}{c}{\emph{transformer-based models}} \\
\hline
DETR (DC5) \cite{carion2020end} & 15.5 & 29.4 &  4.3 & 15.1 & 26.7\\
Dynamic DETR \cite{dai2021dynamic} & 42.9 & 61.0 & 24.6 & 44.9 & 54.4 \\
DN-Deformable-DETR \cite{li2022dn} & 43.4 & 61.9 &  24.8 & 46.8 & 59.4\\
DINO \cite{zhang2022dino} & 49.0 & 66.6 & 32.0 & 52.3 & 63.0\\
\hline
\multicolumn{6}{c}{\emph{CNN-based models}} \\
\hline
Faster RCNN (FPN) \cite{ren2015faster, lin2017feature} & 37.9 & 58.8 & 22.4 & 41.1 & 49.1\\
DenseCLIP \cite{rao2022denseclip} & 40.2 & 63.2 & 26.3 & 44.2 & 51.0\\
HTC \cite{chen2019hybrid} & 42.3 & 61.1 & 23.7 & 45.6 & 56.3 \\
Dyhead \cite{dai2021dynamic} & 43.0 & 60.7 & 24.7 & 46.4 & 53.9\\
R(Det)$^2$ + Cascade \cite{li2022r} & 42.5 & 61.0 & 24.6 & 45.5 & 57.0\\
Softteacher $^\S$ \cite{xu2021end} &  44.5 & - & - & - & -\\
UniDetector (ours) & \textbf{49.3} & \textbf{67.5}&  \textbf{33.3} & \textbf{53.1} & \textbf{63.6} \\
\Xhline{1.1pt} 
\end{tabular}
}
\label{tab:closed}
\end{table}

A universal object detection model should not only generalize well to the open world, but also keep its superiority in the closed world that has been seen during training. We thus train our UniDetector using only images from the COCO training set and evaluate it on the COCO 2017 validation set. We compare our results with existing state-of-the-art closed-world detection models and present the detection AP in Tab. \ref{tab:closed}. In this subsection, we utilize R(Det)$^2$ \cite{li2022r} with the cascade structure \cite{Cai_2018_CVPR} for our detector. For our CLN, we introduce the Dyhead \cite{dai2021dynamic} structure, and focal loss \cite{lin2017focal} for classification. AdamW \cite{kingma2014adam, loshchilov2017decoupled} optimizer is adopted, with 0.00002 for the initial learning rate. 

With the ResNet50 backbone and the 1$\times$ schedule, our UniDetector obtains the 49.3\% AP with a pure CNN-based structure. We surpass the Dyhead \cite{dai2021dynamic}, the state-of-the-art CNN detector by 6.3\% AP. Compared to Softteacher \cite{xu2021end}, a semi-supervised model that utilizes additional images and trains with more epochs, our UniDetector also achieves a 4.8\% higher AP. Compared to recent transformer-based detectors, the performance superiority is also obvious. The results show that our UniDetector not only generalizes well in the open world, but also holds effectiveness in the closed world. The superiority on both the open world and closed world strongly confirms the universality of our UniDetector.


\subsection{Object Detection in the Wild}

\begin{table}[t]
\centering
\setlength{\abovecaptionskip}{0pt}
\setlength{\belowcaptionskip}{0pt}
\caption{\textbf{Zero-shot performance on 13 ODinW datasets.}}
\resizebox{ \columnwidth}{!}{
\begin{tabular}{c|c|c|c}
\Xhline{1.1pt} 
 Model  & \#Data &  Datasets   & Avg. AP \\
\hline
GLIP-T (A) \cite{li2022grounded} & 0.66M & Objects365 & 28.8\\
GLIP-T (B) & 0.66M &  Objects365 & 33.2 \\
GLIP-T (C) &  1.46M &  Objects365, GoldG & 44.4 \\
GLIP-T & 5.46M &  Objects365, GoldG, Cap4M & 46.5 \\
UniDetector (ours) & \textbf{173k} & subset of COCO, Objects365, OpenImages & \textbf{47.3} \\
\Xhline{1.1pt} 
\end{tabular}
}
\label{tab:odinw}
\end{table}

To further demonstrate the ability of our UniDetector to detect everything in every scene, we follow \cite{li2022grounded} to conduct experiments on 13 ODinW datasets. These datasets cover various domains, such as airdrone, underwater, thermal, thus also with a diversity of categories. Such property makes it suitable to measure the universality of a detector. We list the average AP on these 13 datasets in Tab. \ref{tab:odinw}. Compared to GLIP-T, whose backbone (Swin-Tiny) requires a little more budget than ours (ResNet50), our method achieves a higher average AP (47.3\% \emph{v.s.} 46.5\%). In comparison, our method only utilizes 3\% amount of data of GLIP-T. This experiment corroborates the universality of UniDetector and illustrates its excellent data efficiency.

\subsection{Comparison with Open-vocabulary Methods}

\begin{table}[t]
\centering
\setlength{\abovecaptionskip}{0pt}
\setlength{\belowcaptionskip}{0pt}
\caption{\textbf{Comparison with existing open-vocabulary detection methods on the COCO dataset.}}
\resizebox{0.6\columnwidth}{!}{
\begin{tabular}{c|ccc}
\Xhline{1.1pt} 
 Model  & novel & base & all \\
 \hline
 OVR-CNN \cite{zareian2021open} & 22.8 & 46.0 & 39.9 \\
 HierKD \cite{ma2022open} & 20.3 & 51.3 & 43.2 \\
ViLD \cite{gu2021open} & 27.6 & 59.5 & 51.3 \\
RegionCLIP \cite{zhong2022regionclip} & 31.4 & 57.1 & 50.4 \\
OV-DETR \cite{zang2022open} & 29.4 & 61.0 & 52.7\\
PromptDet \cite{feng2022promptdet} & 26.6 & - & 50.6 \\
Detic \cite{zhou2022detecting} & 27.8 & 47.1 & 45.0 \\
UniDetector (ours) & \textbf{35.2}  & 56.8 & 51.2  \\
\Xhline{1.1pt} 
\end{tabular}
}
\label{tab:openvocabulary}
\end{table}


\begin{table}[t]
\centering
\setlength{\abovecaptionskip}{0pt}
\setlength{\belowcaptionskip}{0pt}
\caption{\textbf{Comparison with existing open-vocabulary detection methods on the LVIS v1 dataset.} For LVIS based training, Detic and our method use LVIS base images and image-level annotated images from ImageNet. For unrestricted open-vocabulary training, we only use 10\% amount of images from Objects365.}
\resizebox{ \columnwidth}{!}{
\begin{tabular}{c|c|cc|cc}
\Xhline{1.1pt} 
 Model  &  Data & AP$_r^b$ &  AP$^b$ & AP$_r^m$  & AP$^m$ \\
\hline
\multicolumn{6}{c}{\emph{LVIS based training}} \\
\hline
ViLD \cite{gu2021open} & LVIS base & 16.7 & 27.8 & 16.6 & 25.5\\
RegionCLIP \cite{zhong2022regionclip} & LVIS base & 17.1 & 28.2 & - & -\\
DetPro \cite{du2022learning} & LVIS base & 20.8 & 28.4 & 19.8 & 25.9 \\
OV-DETR \cite{zang2022open} & LVIS base & - & - & 17.3  & 26.6\\
PromptDet \cite{feng2022promptdet} & LVIS base & 21.8 & 27.3 & 21.4 & 25.3\\
Detic \cite{zhou2022detecting} & LVIS base + IN-L & 26.7  & 36.2 & 24.9  & 32.4\\
UniDetector (ours)  & LVIS base + IN-L & \textbf{29.3} & \textbf{36.8} & \textbf{26.5}  & \textbf{32.5}\\
\hline
\multicolumn{6}{c}{\emph{Unrestricted open-vocabulary training}} \\
\hline 
GLIP, Swin-T \cite{li2022grounded} & O365 (all), GlodG, ... & 10.1 & 17.2 & - & - \\
OWL-ViT, ViT-B/32 \cite{minderer2022simple} &  O365 (all), VG & 18.9 & 22.1 & - & -\\
UniDetector (ours), ResNet50 & O365 (10\%), VG & \textbf{20.2} & \textbf{23.4} & - & - \\
\Xhline{1.1pt} 
\end{tabular}
}
\label{tab:openvocabularylvis}
\end{table}

We conduct experiments on the settings of existing open-vocabulary works for a fair comparison with them to further show the effectiveness of our UniDetector,  Specifically, the COCO dataset and the LVIS v1 dataset are splitted in the 48/17 and 866/337 way separately for base and novel classes. For the LVIS experiment, we adopt the same CenterNet2 \cite{zhou2021probabilistic} structure and image-level annotated images as Detic \cite{zhou2022detecting} for detection learning, and Dyhead \cite{dai2021dynamic} for proposal generation. The box and mask AP on novel and base classes is listed in Tab. \ref{tab:openvocabulary} and Tab. \ref{tab:openvocabularylvis}

The obtained box AP powerfully demonstrates the generalization ability of our UniDetector to novel classes. On the COCO dataset, we obtain the 35.2\% box AP for novel classes, which surpasses the best previous method (31.7\% from RegionCLIP) by 3.5\%. On the LVIS dataset, we obtain the 29.3\% box AP and 26.5\% mask AP for novel classes (\emph{i.e.} AP$_r$ in this case), which outperforms Detic by 2.6\% and 1.6\% separately. The extraordinary improvement on novel categories validates the excellent ability of our method for unseen classes. It is worth mentioning that only one detection dataset is involved in this experimental setting, where our UniDetector is even a little restricted by the single source of images. When introducing multiple datasets for training, the superiority of our method is more prominent. With only 10\% amount of the training images, we surpass OWL-ViT by 1.3\% for novel categories. The comparison well demonstrates the universality.


\subsection{Ablation Study}

Finally, we conduct ablation studies in this subsection. We mainly analyze the effect of decoupling region proposal generation and probability calibration here.

\begin{table}[t]
\centering
\setlength{\abovecaptionskip}{0pt}
\setlength{\belowcaptionskip}{0pt}
\caption{\textbf{Ablation study on region proposal generation} on the LVIS v0.5 dataset. The networks listed here are all trained on the subset of COCO and Objects365 dataset. }
\resizebox{\columnwidth}{!}{
\begin{tabular}{c|c|cccc}
\Xhline{1.1pt} 
 decouple  & proposal generation model & AP & AP$_r$  & AP$_c$ & AP$_f$ \\
\hline
 & RPN & 18.1 & 19.0 & 17.6 & 18.9 \\
 \checkmark & RPN  & 19.1 & 19.4 & 18.7 & 18.9\\ 
\checkmark & Faster RCNN (class-agnostic) & 19.7 & 20.4	& 19.3 & 19.9 \\ 
\checkmark & OLN \cite{kim2022learning} & 19.7	& 20.4 & 19.0 & 20.3 \\
 \checkmark & Cascade RPN \cite{vu2019cascade} & 20.0 & 21.5 & 19.1 & 20.5\\ 
\checkmark & CLN (ours) & \textbf{21.2} & \textbf{22.0} & \textbf{20.6} & \textbf{21.0} \\
\Xhline{1.1pt} 
\end{tabular}
}
\label{tab:cln}
\end{table}

\paragraph{Decoupling proposal generation and RoI classification.} Tab. \ref{tab:cln} analyzes the effect of the decoupling training manner. A trivial Faster RCNN trained on COCO and Objects365 obtains the 18.1\% open-world AP on LVIS. If we decouple the two stages, the box AP is 19.1\%. The 1.0\% AP improvement demonstrates that the decoupling manner is beneficial for open world detection, while this does not happen in traditional closed world detection. If we extract region proposals with a class-agnostic Faster RCNN, the AP is 19.7\%. The 0.6\% improvement indicates that the structure with both RPN and RoI head is more suitable for generating proposals in the open world than a single RPN. If we adopt the OLN \cite{kim2022learning}, also with the RoI head, the LVIS AP is still 19.7\%, which indicates that pure localization information cannot bring a further improvement. Our CLN, with both classification score and localization quality, contributes to a 21.2\% AP. This AP is higher than not only networks with similar budgets, but also more complicated models like Cascade RPN. This demonstrates the effectiveness of the decoupling learning manner and our CLN.

\begin{figure}[t]
\centering
\setlength{\abovecaptionskip}{0pt}
\setlength{\belowcaptionskip}{0pt}
\includegraphics[width=\columnwidth]{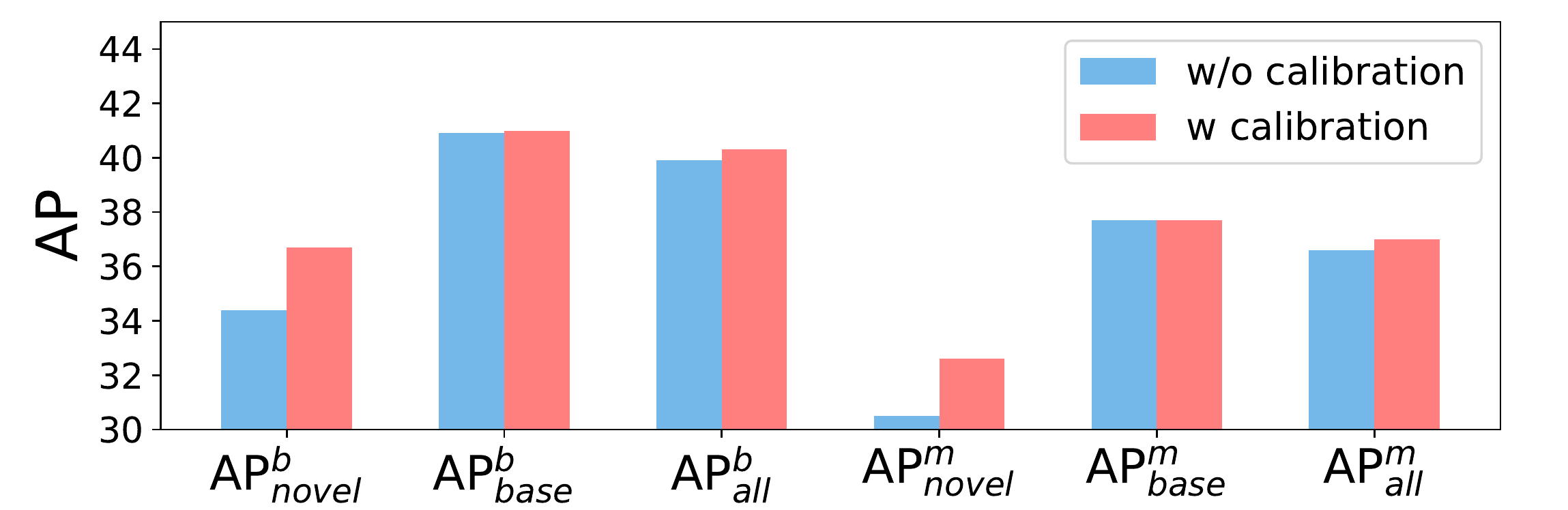}
\caption{\textbf{Illustration for probability calibration} on the LVIS v0.5 dataset. We train the Centernet2 model on LVIS base images and image-level annotated images from ImageNet.}
\label{fig:calib}
\end{figure}

\paragraph{Probability calibration} We further measure the AP on novel and base categories separately to test the ability of probability calibration. We follow the settings of Detic \cite{zhou2022detecting} on LVIS v0.5 and plot the box and mask AP in Fig. \ref{fig:calib}. We notice that after calibration, both box AP and mask AP on novel classes improve significantly, more than 2\%. As a result, the performance gap between base and novel classes is reduced remarkably. In comparison, the performance on base classes almost remains the same. This is because the prior probability we design reduces the self-confidence of base categories significantly. As we can see, the probability calibration alleviates the bias of trained models, thus helping generate more balanced predictions in the open world.

\section{Conclusion}

In this paper, we propose UniDetector, a universal object detection framework. By utilizing images of multiple sources, heterogeneous label spaces, and generalizing the detector to the open world, our UniDetector can directly detect everything in every scene without any finetuning. Extensive experiments on large-vocabulary datasets and diverse scenes demonstrate its strong universality - it behaves the ability to identify the most categories so far. Universality is a vital issue that bridges the gap between artificial intelligence systems and biological mechanisms. We believe our research will stimulate following research along the universal computer vision direction in the future. 



{\small
\bibliographystyle{ieee_fullname}
\bibliography{egbib}

\begin{thebibliography}{10}\itemsep=-1pt

\bibitem{bansal2018zero}
Ankan Bansal, Karan Sikka, Gaurav Sharma, Rama Chellappa, and Ajay Divakaran.
\newblock Zero-shot object detection.
\newblock In {\em ECCV}, 2018.

\bibitem{bochkovskiy2020yolov4}
Alexey Bochkovskiy, Chien-Yao Wang, and Hong-Yuan~Mark Liao.
\newblock Yolov4: Optimal speed and accuracy of object detection.
\newblock {\em arXiv:2004.10934}, 2020.

\bibitem{cai2022bigdetection}
Likun Cai, Zhi Zhang, Yi Zhu, Li Zhang, Mu Li, and Xiangyang Xue.
\newblock Bigdetection: A large-scale benchmark for improved object detector
  pre-training.
\newblock In {\em CVPR}, 2022.

\bibitem{Cai_2018_CVPR}
Zhaowei Cai and Nuno Vasconcelos.
\newblock Cascade r-cnn: Delving into high quality object detection.
\newblock In {\em CVPR}, 2018.

\bibitem{carion2020end}
Nicolas Carion, Francisco Massa, Gabriel Synnaeve, Nicolas Usunier, Alexander
  Kirillov, and Sergey Zagoruyko.
\newblock End-to-end object detection with transformers.
\newblock In {\em ECCV}, 2020.

\bibitem{chen2019hybrid}
Kai Chen, Jiangmiao Pang, Jiaqi Wang, Yu Xiong, Xiaoxiao Li, Shuyang Sun,
  Wansen Feng, Ziwei Liu, Jianping Shi, Wanli Ouyang, et~al.
\newblock Hybrid task cascade for instance segmentation.
\newblock In {\em CVPR}, 2019.

\bibitem{chen2019mmdetection}
Kai Chen, Jiaqi Wang, Jiangmiao Pang, Yuhang Cao, Yu Xiong, Xiaoxiao Li,
  Shuyang Sun, Wansen Feng, Ziwei Liu, Jiarui Xu, et~al.
\newblock Mmdetection: Open mmlab detection toolbox and benchmark.
\newblock {\em arXiv:1906.07155}, 2019.

\bibitem{cordts2016cityscapes}
Marius Cordts, Mohamed Omran, Sebastian Ramos, Timo Rehfeld, Markus Enzweiler,
  Rodrigo Benenson, Uwe Franke, Stefan Roth, and Bernt Schiele.
\newblock The cityscapes dataset for semantic urban scene understanding.
\newblock In {\em CVPR}, 2016.

\bibitem{dai2021dynamic}
Xiyang Dai, Yinpeng Chen, Bin Xiao, Dongdong Chen, Mengchen Liu, Lu Yuan, and
  Lei Zhang.
\newblock Dynamic head: Unifying object detection heads with attentions.
\newblock In {\em CVPR}, 2021.

\bibitem{dai2021up}
Zhigang Dai, Bolun Cai, Yugeng Lin, and Junying Chen.
\newblock Up-detr: Unsupervised pre-training for object detection with
  transformers.
\newblock In {\em CVPR}, 2021.

\bibitem{du2022learning}
Yu Du, Fangyun Wei, Zihe Zhang, Miaojing Shi, Yue Gao, and Guoqi Li.
\newblock Learning to prompt for open-vocabulary object detection with
  vision-language model.
\newblock In {\em CVPR}, 2022.

\bibitem{everingham2010pascal}
Mark Everingham, Luc Van~Gool, Christopher~KI Williams, John Winn, and Andrew
  Zisserman.
\newblock The pascal visual object classes (voc) challenge.
\newblock {\em IJCV}, 2010.

\bibitem{feng2022promptdet}
Chengjian Feng, Yujie Zhong, Zequn Jie, Xiangxiang Chu, Haibing Ren, Xiaolin
  Wei, Weidi Xie, and Lin Ma.
\newblock Promptdet: Towards open-vocabulary detection using uncurated images.
\newblock In {\em ECCV}, 2022.

\bibitem{girshick2015fast}
Ross Girshick.
\newblock Fast r-cnn.
\newblock In {\em ICCV}, 2015.

\bibitem{girshick2014rich}
Ross Girshick, Jeff Donahue, Trevor Darrell, and Jitendra Malik.
\newblock Rich feature hierarchies for accurate object detection and semantic
  segmentation.
\newblock In {\em CVPR}, 2014.

\bibitem{gu2021open}
Xiuye Gu, Tsung-Yi Lin, Weicheng Kuo, and Yin Cui.
\newblock Open-vocabulary object detection via vision and language knowledge
  distillation.
\newblock {\em arXiv:2104.13921}, 2021.

\bibitem{gupta2019lvis}
Agrim Gupta, Piotr Dollar, and Ross Girshick.
\newblock Lvis: A dataset for large vocabulary instance segmentation.
\newblock In {\em CVPR}, 2019.

\bibitem{he2017mask}
Kaiming He, Georgia Gkioxari, Piotr Doll{\'a}r, and Ross Girshick.
\newblock Mask r-cnn.
\newblock In {\em ICCV}, 2017.

\bibitem{he2016deep}
Kaiming He, Xiangyu Zhang, Shaoqing Ren, and Jian Sun.
\newblock Deep residual learning for image recognition.
\newblock In {\em CVPR}, 2016.

\bibitem{jia2021scaling}
Chao Jia, Yinfei Yang, Ye Xia, Yi-Ting Chen, Zarana Parekh, Hieu Pham, Quoc Le,
  Yun-Hsuan Sung, Zhen Li, and Tom Duerig.
\newblock Scaling up visual and vision-language representation learning with
  noisy text supervision.
\newblock In {\em ICML}, 2021.

\bibitem{kim2022learning}
Dahun Kim, Tsung-Yi Lin, Anelia Angelova, In~So Kweon, and Weicheng Kuo.
\newblock Learning open-world object proposals without learning to classify.
\newblock {\em RAL}, 2022.

\bibitem{kingma2014adam}
Diederik~P Kingma and Jimmy Ba.
\newblock Adam: A method for stochastic optimization.
\newblock {\em arXiv:1412.6980}, 2014.

\bibitem{krishna2017visual}
Ranjay Krishna, Yuke Zhu, Oliver Groth, Justin Johnson, Kenji Hata, Joshua
  Kravitz, Stephanie Chen, Yannis Kalantidis, Li-Jia Li, David~A Shamma, et~al.
\newblock Visual genome: Connecting language and vision using crowdsourced
  dense image annotations.
\newblock {\em IJCV}, 2017.

\bibitem{NIPS2012_c399862d}
Alex Krizhevsky, Ilya Sutskever, and Geoffrey~E Hinton.
\newblock Imagenet classification with deep convolutional neural networks.
\newblock In {\em NeurIPS}, 2012.

\bibitem{kuznetsova2020open}
Alina Kuznetsova, Hassan Rom, Neil Alldrin, Jasper Uijlings, Ivan Krasin, Jordi
  Pont-Tuset, Shahab Kamali, Stefan Popov, Matteo Malloci, Alexander
  Kolesnikov, et~al.
\newblock The open images dataset v4.
\newblock {\em IJCV}, 2020.

\bibitem{law2018cornernet}
Hei Law and Jia Deng.
\newblock Cornernet: Detecting objects as paired keypoints.
\newblock In {\em ECCV}, 2018.

\bibitem{li2022dn}
Feng Li, Hao Zhang, Shilong Liu, Jian Guo, Lionel~M Ni, and Lei Zhang.
\newblock Dn-detr: Accelerate detr training by introducing query denoising.
\newblock In {\em CVPR}, 2022.

\bibitem{li2022grounded}
Liunian~Harold Li, Pengchuan Zhang, Haotian Zhang, Jianwei Yang, Chunyuan Li,
  Yiwu Zhong, Lijuan Wang, Lu Yuan, Lei Zhang, Jenq-Neng Hwang, et~al.
\newblock Grounded language-image pre-training.
\newblock In {\em CVPR}, 2022.

\bibitem{li2022r}
Yali Li and Shengjin Wang.
\newblock R (det) 2: Randomized decision routing for object detection.
\newblock In {\em CVPR}, 2022.

\bibitem{lin2017feature}
Tsung-Yi Lin, Piotr Doll{\'a}r, Ross Girshick, Kaiming He, Bharath Hariharan,
  and Serge Belongie.
\newblock Feature pyramid networks for object detection.
\newblock In {\em CVPR}, 2017.

\bibitem{lin2017focal}
Tsung-Yi Lin, Priya Goyal, Ross Girshick, Kaiming He, and Piotr Doll{\'a}r.
\newblock Focal loss for dense object detection.
\newblock In {\em ICCV}, 2017.

\bibitem{lin2014microsoft}
Tsung-Yi Lin, Michael Maire, Serge Belongie, James Hays, Pietro Perona, Deva
  Ramanan, Piotr Doll{\'a}r, and C~Lawrence Zitnick.
\newblock Microsoft coco: Common objects in context.
\newblock In {\em ECCV}, 2014.

\bibitem{liu2016ssd}
Wei Liu, Dragomir Anguelov, Dumitru Erhan, Christian Szegedy, Scott Reed,
  Cheng-Yang Fu, and Alexander~C Berg.
\newblock Ssd: Single shot multibox detector.
\newblock In {\em ECCV}, 2016.

\bibitem{loshchilov2017decoupled}
Ilya Loshchilov and Frank Hutter.
\newblock Decoupled weight decay regularization.
\newblock {\em arXiv:1711.05101}, 2017.

\bibitem{ma2022open}
Zongyang Ma, Guan Luo, Jin Gao, Liang Li, Yuxin Chen, Shaoru Wang, Congxuan
  Zhang, and Weiming Hu.
\newblock Open-vocabulary one-stage detection with hierarchical visual-language
  knowledge distillation.
\newblock In {\em CVPR}, 2022.

\bibitem{meng2022detection}
Lingchen Meng, Xiyang Dai, Yinpeng Chen, Pengchuan Zhang, Dongdong Chen,
  Mengchen Liu, Jianfeng Wang, Zuxuan Wu, Lu Yuan, and Yu-Gang Jiang.
\newblock Detection hub: Unifying object detection datasets via query
  adaptation on language embedding.
\newblock {\em arXiv:2206.03484}, 2022.

\bibitem{minderer2022simple}
Matthias Minderer, Alexey Gritsenko, Austin Stone, Maxim Neumann, Dirk
  Weissenborn, Alexey Dosovitskiy, Aravindh Mahendran, Anurag Arnab, Mostafa
  Dehghani, Zhuoran Shen, et~al.
\newblock Simple open-vocabulary object detection with vision transformers.
\newblock In {\em ECCV}, 2022.

\bibitem{peng2020large}
Junran Peng, Xingyuan Bu, Ming Sun, Zhaoxiang Zhang, Tieniu Tan, and Junjie
  Yan.
\newblock Large-scale object detection in the wild from imbalanced
  multi-labels.
\newblock In {\em CVPR}, 2020.

\bibitem{radford2021learning}
Alec Radford, Jong~Wook Kim, Chris Hallacy, Aditya Ramesh, Gabriel Goh,
  Sandhini Agarwal, Girish Sastry, Amanda Askell, Pamela Mishkin, Jack Clark,
  et~al.
\newblock Learning transferable visual models from natural language
  supervision.
\newblock In {\em ICML}, 2021.

\bibitem{rahman2020improved}
Shafin Rahman, Salman Khan, and Nick Barnes.
\newblock Improved visual-semantic alignment for zero-shot object detection.
\newblock In {\em AAAI}, 2020.

\bibitem{rao2022denseclip}
Yongming Rao, Wenliang Zhao, Guangyi Chen, Yansong Tang, Zheng Zhu, Guan Huang,
  Jie Zhou, and Jiwen Lu.
\newblock Denseclip: Language-guided dense prediction with context-aware
  prompting.
\newblock In {\em CVPR}, 2022.

\bibitem{redmon2016you}
Joseph Redmon, Santosh Divvala, Ross Girshick, and Ali Farhadi.
\newblock You only look once: Unified, real-time object detection.
\newblock In {\em CVPR}, 2016.

\bibitem{ren2015faster}
Shaoqing Ren, Kaiming He, Ross Girshick, and Jian Sun.
\newblock Faster r-cnn: Towards real-time object detection with region proposal
  networks.
\newblock In {\em NeurIPS}, 2015.

\bibitem{russakovsky2015imagenet}
Olga Russakovsky, Jia Deng, Hao Su, Jonathan Krause, Sanjeev Satheesh, Sean Ma,
  Zhiheng Huang, Andrej Karpathy, Aditya Khosla, Michael Bernstein, et~al.
\newblock Imagenet large scale visual recognition challenge.
\newblock {\em IJCV}, 2015.

\bibitem{shao2019objects365}
Shuai Shao, Zeming Li, Tianyuan Zhang, Chao Peng, Gang Yu, Xiangyu Zhang, Jing
  Li, and Jian Sun.
\newblock Objects365: A large-scale, high-quality dataset for object detection.
\newblock In {\em ICCV}, 2019.

\bibitem{shi2021multi}
Bowen Shi, Xiaopeng Zhang, Haohang Xu, Wenrui Dai, Junni Zou, Hongkai Xiong,
  and Qi Tian.
\newblock Multi-dataset pretraining: A unified model for semantic segmentation.
\newblock {\em arXiv:2106.04121}, 2021.

\bibitem{solovyev2021weighted}
Roman Solovyev, Weimin Wang, and Tatiana Gabruseva.
\newblock Weighted boxes fusion: Ensembling boxes from different object
  detection models.
\newblock {\em Image and Vision Computing}, 2021.

\bibitem{tan2021equalization}
Jingru Tan, Xin Lu, Gang Zhang, Changqing Yin, and Quanquan Li.
\newblock Equalization loss v2: A new gradient balance approach for long-tailed
  object detection.
\newblock In {\em CVPR}, 2021.

\bibitem{tan2020equalization}
Jingru Tan, Changbao Wang, Buyu Li, Quanquan Li, Wanli Ouyang, Changqing Yin,
  and Junjie Yan.
\newblock Equalization loss for long-tailed object recognition.
\newblock In {\em CVPR}, 2020.

\bibitem{tian2019fcos}
Zhi Tian, Chunhua Shen, Hao Chen, and Tong He.
\newblock Fcos: Fully convolutional one-stage object detection.
\newblock In {\em ICCV}, 2019.

\bibitem{vu2019cascade}
Thang Vu, Hyunjun Jang, Trung~X Pham, and Chang Yoo.
\newblock Cascade rpn: Delving into high-quality region proposal network with
  adaptive convolution.
\newblock {\em NeurIPS}, 2019.

\bibitem{wang2021seesaw}
Jiaqi Wang, Wenwei Zhang, Yuhang Zang, Yuhang Cao, Jiangmiao Pang, Tao Gong,
  Kai Chen, Ziwei Liu, Chen~Change Loy, and Dahua Lin.
\newblock Seesaw loss for long-tailed instance segmentation.
\newblock In {\em CVPR}, 2021.

\bibitem{wang2019towards}
Xudong Wang, Zhaowei Cai, Dashan Gao, and Nuno Vasconcelos.
\newblock Towards universal object detection by domain attention.
\newblock In {\em CVPR}, 2019.

\bibitem{xu2021end}
Mengde Xu, Zheng Zhang, Han Hu, Jianfeng Wang, Lijuan Wang, Fangyun Wei, Xiang
  Bai, and Zicheng Liu.
\newblock End-to-end semi-supervised object detection with soft teacher.
\newblock In {\em ICCV}, 2021.

\bibitem{yu2020bdd100k}
Fisher Yu, Haofeng Chen, Xin Wang, Wenqi Xian, Yingying Chen, Fangchen Liu,
  Vashisht Madhavan, and Trevor Darrell.
\newblock Bdd100k: A diverse driving dataset for heterogeneous multitask
  learning.
\newblock In {\em CVPR}, 2020.

\bibitem{zang2022open}
Yuhang Zang, Wei Li, Kaiyang Zhou, Chen Huang, and Chen~Change Loy.
\newblock Open-vocabulary detr with conditional matching.
\newblock In {\em ECCV}, 2022.

\bibitem{zareian2021open}
Alireza Zareian, Kevin~Dela Rosa, Derek~Hao Hu, and Shih-Fu Chang.
\newblock Open-vocabulary object detection using captions.
\newblock In {\em CVPR}, 2021.

\bibitem{zhai2022lit}
Xiaohua Zhai, Xiao Wang, Basil Mustafa, Andreas Steiner, Daniel Keysers,
  Alexander Kolesnikov, and Lucas Beyer.
\newblock Lit: Zero-shot transfer with locked-image text tuning.
\newblock In {\em CVPR}, 2022.

\bibitem{zhang2017mixup}
Hongyi Zhang, Moustapha Cisse, Yann~N Dauphin, and David Lopez-Paz.
\newblock mixup: Beyond empirical risk minimization.
\newblock {\em arXiv:1710.09412}, 2017.

\bibitem{zhang2022dino}
Hao Zhang, Feng Li, Shilong Liu, Lei Zhang, Hang Su, Jun Zhu, Lionel~M Ni, and
  Heung-Yeung Shum.
\newblock Dino: Detr with improved denoising anchor boxes for end-to-end object
  detection.
\newblock {\em arXiv:2203.03605}, 2022.

\bibitem{zhang2020bridging}
Shifeng Zhang, Cheng Chi, Yongqiang Yao, Zhen Lei, and Stan~Z Li.
\newblock Bridging the gap between anchor-based and anchor-free detection via
  adaptive training sample selection.
\newblock In {\em CVPR}, 2020.

\bibitem{zhao2020object}
Xiangyun Zhao, Samuel Schulter, Gaurav Sharma, Yi-Hsuan Tsai, Manmohan
  Chandraker, and Ying Wu.
\newblock Object detection with a unified label space from multiple datasets.
\newblock In {\em ECCV}, 2020.

\bibitem{zhong2022regionclip}
Yiwu Zhong, Jianwei Yang, Pengchuan Zhang, Chunyuan Li, Noel Codella,
  Liunian~Harold Li, Luowei Zhou, Xiyang Dai, Lu Yuan, Yin Li, et~al.
\newblock Regionclip: Region-based language-image pretraining.
\newblock In {\em CVPR}, 2022.

\bibitem{zhou2022detecting}
Xingyi Zhou, Rohit Girdhar, Armand Joulin, Phillip Kr{\"a}henb{\"u}hl, and
  Ishan Misra.
\newblock Detecting twenty-thousand classes using image-level supervision.
\newblock {\em arXiv:2201.02605}, 2022.

\bibitem{zhou2021probabilistic}
Xingyi Zhou, Vladlen Koltun, and Philipp Kr{\"a}henb{\"u}hl.
\newblock Probabilistic two-stage detection.
\newblock {\em arXiv:2103.07461}, 2021.

\bibitem{zhou2022simple}
Xingyi Zhou, Vladlen Koltun, and Philipp Kr{\"a}henb{\"u}hl.
\newblock Simple multi-dataset detection.
\newblock In {\em CVPR}, 2022.

\bibitem{zhou2019objects}
Xingyi Zhou, Dequan Wang, and Philipp Kr{\"a}henb{\"u}hl.
\newblock Objects as points.
\newblock {\em arXiv:1904.07850}, 2019.

\bibitem{zhu2019zero}
Pengkai Zhu, Hanxiao Wang, and Venkatesh Saligrama.
\newblock Zero shot detection.
\newblock {\em TCSVT}, 2019.

\bibitem{zhu2020don}
Pengkai Zhu, Hanxiao Wang, and Venkatesh Saligrama.
\newblock Don't even look once: Synthesizing features for zero-shot detection.
\newblock In {\em CVPR}, 2020.

\bibitem{zhu2018visdrone}
Pengfei Zhu, Longyin Wen, Dawei Du, Xiao Bian, Haibin Ling, Qinghua Hu, Qinqin
  Nie, Hao Cheng, Chenfeng Liu, Xiaoyu Liu, et~al.
\newblock Visdrone-det2018: The vision meets drone object detection in image
  challenge results.
\newblock In {\em ECCVW}, 2018.

\bibitem{zhu2020deformable}
Xizhou Zhu, Weijie Su, Lewei Lu, Bin Li, Xiaogang Wang, and Jifeng Dai.
\newblock Deformable detr: Deformable transformers for end-to-end object
  detection.
\newblock In {\em ICLR}, 2021.

\end{thebibliography}
}

\end{document}